\documentclass[11pt,a4paper]{article}
\usepackage{authblk}
\usepackage[hyperref]{acl2020}
\usepackage{times}
\usepackage{latexsym}

\usepackage{graphicx}
\graphicspath{{./}}
\usepackage{amsfonts} 

\usepackage{microtype}

\aclfinalcopy 

\title{An Attention Ensemble Approach for Efficient Text Classification of Indian Languages}

\author[1]{\textbf{Atharva Kulkarni}}
\author[1]{\textbf{Amey Hengle}}
\author[2]{\textbf{Rutuja Udyawar}}
\affil[1]{Department of Computer Engineering, PVG's COET, Savitribai Phule Pune University, India.}
\affil[2]{Optimum Data Analytics, India.}
\affil[1]{\texttt {\{atharva.j.kulkarni1998, ameyhengle22\}@gmail.com}}
\affil[2]{\texttt {rutuja.udyawar@odaml.com}}

\date{}

\begin{document}
\maketitle
\begin{abstract}
The recent surge of complex attention-based deep learning architectures has led to extraordinary results in various downstream NLP tasks in the English language. However, such research for resource-constrained and morphologically rich Indian vernacular languages has been relatively limited. This paper proffers team SPPU\_AKAH's solution for the TechDOfication 2020 subtask-1f: which focuses on the coarse-grained technical domain identification of short text documents in Marathi, a Devanagari script-based Indian language. Availing the large dataset at hand, a hybrid CNN-BiLSTM attention ensemble model is proposed that competently combines the intermediate sentence representations generated by the convolutional neural network and the bidirectional long short-term memory, leading to efficient text classification. Experimental results show that the proposed model outperforms various baseline machine learning and deep learning models in the given task, giving the best validation accuracy of 89.57\% and f1-score of 0.8875. Furthermore, the solution resulted in the best system submission for this subtask, giving a test accuracy of 64.26\% and f1-score of 0.6157, transcending the performances of other teams as well as the baseline system given by the organizers of the shared task.
\end{abstract}

\section{Introduction}
The advent of attention-based neural networks and the availability of large labelled datasets has resulted in great success and state-of-the-art performance for English text classification \citep{yang2016hierarchical,zhou2016attention,wang2016attention, gao2018hierarchical}. Such results, however, for Indian language text classification tasks are far and few as most of the research employ traditional machine learning and deep learning models \citep{joshi2019deep, tummalapalli2018towards, bolaj2016survey, bolaj2016text, dhar2018performance}. Apart from being heavily consumed in the print format, the growth in the Indian languages internet user base is monumental, scaling from 234 million in 2016 to 536 million by 2021 \footnote{https://home.kpmg/in/en/home/insights/2017/04/indian-language-internet-users.html}. Even so, just like most other Indian languages, the progress in NLP for Marathi has been relatively constrained, due to factors such as the unavailability of large-scale training resources, structural un-similarity with the English language, and a profusion of morphological variations, thus, making the generalization of deep learning architectures to languages like Marathi difficult.

This work posits a solution for the TechDOfication 2020 subtask-1f: coarse-grained domain classification for short Marathi texts. The task provides a large corpus of Marathi text documents spanning across four domains: Biochemistry, Communication Technology, Computer Science, and Physics. Efficient domain identification can potentially impact, and improve research in downstream NLP applications such as question answering, transliteration, machine translation, and text summarization, to name a few. Inspired by the works of \citep{er2016attention, guo2018cran, zheng2019hybrid}, a hybrid CNN-BiLSTM attention ensemble model is proposed in this work. In recent years, Convolutional Neural Networks \citep{kim2014convolutional, conneau2016very, zhang2015character, liu2020multichannel, le2017convolutional} and Recurrent Neural Networks \citep{liu2016recurrent,sundermeyer2015feedforward} have been used quite frequently for text classification tasks. Quite different from one another, the CNNs and the RNNs show different capabilities to generate intermediate text representation. CNN models an input sentence by utilizing convolutional filters to identify the most influential n-grams of different semantic aspects \citep{conneau2016very}. RNN can handle variable-length input sentences and is particularly well suited for modeling sequential data, learning important temporal features and long-term dependencies for robust text representation \citep{hochreiter1997long}. However, whilst CNN can only capture local patterns and fails to incorporate the long-term dependencies and the sequential features, RNN cannot distinguish between keywords that contribute more context to the classification task from the normal stopwords. Thus, the proposed model hypothesizes a potent way to subsume the advantages of both the CNN and the RNN using the attention mechanism. The model employs a parallel structure where both the CNN and the BiLSTM model the input sentences independently. The intermediate representations, thus generated, are combined using the attention mechanism. Therefore, the generated vector has useful temporal features from the sequences generated by the RNN according to the context generated by the CNN. Results attest that the proposed model outperforms various baseline machine learning and deep learning models in the given task, giving the best validation accuracy and f1-score.

\section{Related Work}
Since the past decade, the research in NLP has shifted from a traditional statistical standpoint to complex neural network architectures. The CNN and RNN based architectures have emerged greatly successful for the text classification task. Yoon Kim was the first one who applied a CNN model for English text classification. In this work, a series of experiments were conducted with single as well as multi-channel convolutional neural networks, built on top of randomly generated, pretrained, and fine-tuned word vectors \citep{kim2014convolutional}. This success of CNN for text classification led to the emergence of more complex CNN models \citep{conneau2016very} as well as CNN models with character level inputs \citep{zhang2015character}. RNNs are capable of generating effective text representation by learning temporal features and long-term dependencies between the words \citep{hochreiter1997long,graves2005framewise}. However, these methods treat each word in the sentences equally and thus cannot distinguish between the keywords that contribute more to the classification and the common words. Hybrid models proposed by \citep{xiao2016efficient} and \citep{8260793} succeed in exploiting the advantages of both CNN  and RNN, by using them in combination for text classification.

Since the introduction of the attention mechanism \citep{vaswani2017attention}, it has become an effective strategy for dynamically learning the contribution of different features to specific tasks. Needless to say, the attention mechanism has expeditiously found its way into NLP literature, with many works effectively leveraging it to improve the text classification task. \citep{guo2018cran} proposed a CNN - RNN attention-based neural network (CRAN) for text classification. This work illustrates the effectiveness of using the CNN layer as a context of the attention model. Results show that using this mechanism enables the proposed model to pick the important words from the sequences generated by the RNN layer, thus helping it to outperform many baselines as well as hybrid attention-based models in the text classification task. \citep{er2016attention} proposed an attention pooling strategy, which focuses on making a model learn better sentence representations for improved text classification. Authors use the intermediate sentence representations produced by a BiLSTM layer in reference with the local representations produced by a CNN layer to obtain the attention weights.  Experimental results demonstrate that the proposed model outperforms state-of-the-art approaches on a number of benchmark datasets for text classification. \citep{zheng2019hybrid} combine the BiLSTM and CNN with the attention mechanism for fine-grained text classification tasks. The authors employ a method in which intermediate sentence representations generated by BiLSTM are passed to a CNN layer which is then max pooled to capture the local features of a sentence. The local feature representations are further combined by using an attention layer to calculate the attention weights. In this way, the attention layer can assign different weights to features according to their importance to the text classification task.

The literature in NLP focusing on the resource-constrained Indian languages has been fairly restrained. \citep{tummalapalli2018towards} evaluated the performance of vanilla CNN, LSTM, and multi-Input CNN for the text-classification of Hindi and Telugu texts. The results indicate that CNN based models perform surprisingly better as compared to LSTM and SVM using n-gram features. \citep{joshi2019deep} have compared different deep learning approaches for Hindi sentence classification. The authors have evaluated the effect of using pretrained fasttext Hindi embeddings on the sentence classification task. The finest classification performance is achieved by the Vanilla CNN model when initialized with fasttext word embeddings fine-tuned on the specific dataset.

\section{Dataset}
The TechDOfication-2020 subtask-1f dataset consists of labelled Marathi text documents, each belonging to one of the four classes, namely: Biochemistry (bioche), Communication Technology (com\_tech), Computer Science (cse), and Physics (phy). The training data has a mean length of 26.89 words with a standard deviation of 25.12.

Table~\ref{data-distribution} provides an overview of the distribution of the corpus across the four labels for training and validation data. From the table, it is evident that the dataset is imbalanced, with the class Communication Technology and Biochemistry having the most and the least documents, respectively. It is, therefore, reasonable to postulate that this data imbalance may lead to the overfitting of the model on some classes. This is further articulated in the Results section.

\begin{table}
\centering
\begin{tabular}{lll}
\hline \textbf{Label} & \textbf{Training Data} & \textbf{Validation Data} \\ \hline
bioche & 5,002 & 420 \\
com\_tech & 17,995 & 1,505 \\
cse & 9,344 & 885 \\
phy & 9,656 & 970 \\
Total & 41,997 & 3,780 \\
\hline
\end{tabular}
\caption{\label{data-distribution} Data distribution.}
\end{table}

\section{Proposed Model}
This section describes the proposed multi-input attention-based parallel CNN-BiLSTM. Figure~\ref{Architecture} depicts the model architecture. Each component is explained in detail as follows:

\begin{figure}[ht]
    \includegraphics[width =  7.5 cm, height= 12 cm]{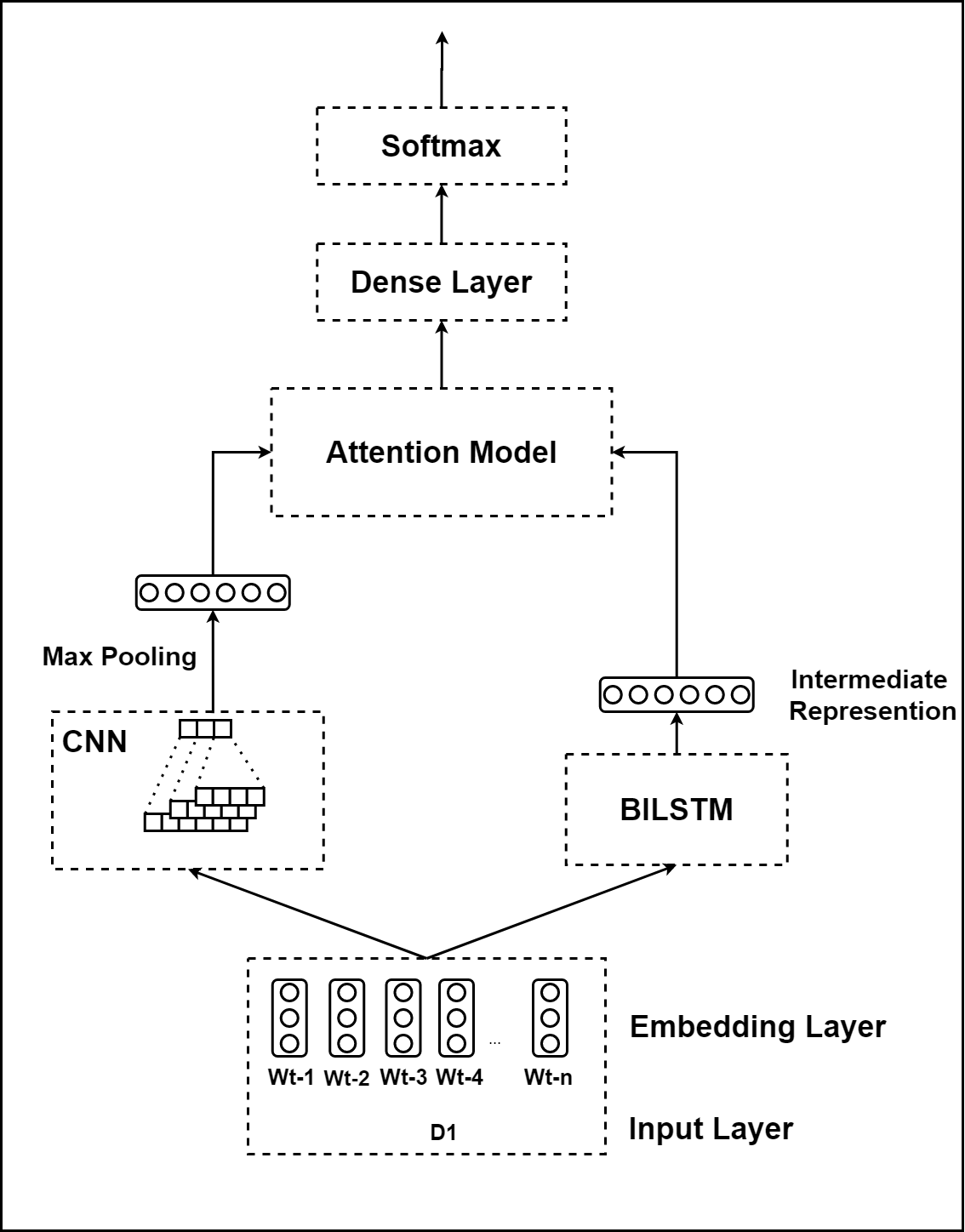}
\caption{\label{Architecture}Model Architecture.}
\end{figure}

\subsection{Word Embedding Layer}
The proposed model uses fasttext word embeddings trained on the unsupervised skip-gram model to map the words from the corpus vocabulary to a corresponding dense vector. Fasttext embeddings are preferred over the word2vec \citep{mikolov2013efficient} or glove variants \citep{pennington2014glove}, as fasttext represents each word as a sequence of character-n-grams, which in turn helps to capture the morphological richness of languages like Marathi. The embedding layer converts each word $w_i$ in the document $T=\{w_1, w_2, ..., w_n\}$ of $n$ words, into a real-valued dense vector $e_i$ using the following matrix-vector product:
\begin{equation}
\label{word-embedding-equation}
e_i = Wv_i
\end{equation}
where $W \in \mathbb{R}^{d\times|V|}$ is the embedding matrix, $|V|$ is a fixed-sized vocabulary of the corpus and $d$ is the word embedding dimension. $v_i$ is the one-hot encoded vector with the element $e_i$ set to 1 while the other elements set to 0. Thus, the document can be represented as real-valued vector $e=\{e_1, e_2,..., e_n\}$.

\subsection{Bi-LSTM Layer}
The word embeddings generated by the embeddings layer are fed to the BiLSTM unit step by step. A Bidirectional Long-short term memory (Bi-LSTM) \citep{graves2005framewise} layer is just a combination of two LSTMs \citep{hochreiter1997long} running in opposite directions. This allows the networks to have both forward and backward information about the sequence at every time step, resulting in better understanding and preservation of the context. It is also able to counter the problem of vanishing gradients to a certain extent by utilizing the input, the output, and the forget gates. The intermediate sentence representation generated by Bi-LSTM is denoted as $h$.

\subsection{CNN Layer}
The discrete convolutions performed by the CNN layer on the input word embeddings, help to extract the most influential n-grams in the sentence. Three parallel convolutional layers with three different window sizes are used so that the model can learn multiple types of embeddings of local regions, and complement one another. Finally, the sentence representations of all the different convolutions are concatenated and max-pooled to get the most dominant features. The output is denoted as $c$.

\subsection{Attention Layer}
The linchpin of the model is the attention block that effectively combines the intermediate sentence feature representation generated by BiLSTM with the local feature representation generated by CNN. At each time step $t$, taking the output $h_t$ of the BiLSTM, and $c_t$ of the CNN, the attention weights $\alpha_t$ are calculated as:
\begin{equation}
\label{attention-weights-equation}
u_t = tanh(W_1h_t + W_2c_t + b)
\end{equation}
\begin{equation}
\alpha_t = Softmax(u_t)
\end{equation}
Where $W_1$ and $W_2$ are the attention weights, and $b$ is the attention bias learned via backpropagation. The final sentence representation $s$ is calculated as the weighted arithmetic mean based on the weights $\alpha = \{\alpha_1, \alpha_2, ..., \alpha_n\}$, and output of the BiLSTM $h=\{h_1, h_2, ..., h_n\}$. It is given as:
\begin{equation}
s=\sum_{t=1}^n \alpha_t*h_t
\end{equation}
Thus, the model is able to retain the merits of both the BiLSTM and the CNN, leading to a more robust sentence representation. This representation is then fed to a fully connected layer for dimensionality reduction.

\subsection{Classification Layer}
The output of the fully connected attention layer is passed to a dense layer with softmax activation to predict a discrete label out of the four labels in the given task.

\section{Experimental Setup}
Each text document is tokenized and padded to a maximum length of 100. Longer documents are truncated. The work of \citep{kakwani2020inlpsuite} is referred for selecting the optimal set of hyperparameters for training the fasttext skip-gram model. The 300-dimensional fasttext word embeddings are trained on the given corpus for 50 epochs, with a minimum token count of 1, and 10 negative examples, sampled for each instance. The rest of the hyperparameter values were chosen as default \citep{bojanowski2017enriching}. After training, an average loss of 0.6338. was obtained over 50 epochs. The validation dataset is used to tune the hyperparameters. The LSTM layer dimension is set to 128 neurons with a dropout rate of 0.3. Thus, the BiLSTM gives an intermediate representation of 256 dimensions. For the CNN block, we employ three parallel convolutional layers with filter sizes 3, 4, and 5, each having 256 feature maps. A dropout rate of 0.3 is applied to each layer. The local representations, thus, generated by the parallel CNNs are then concatenated and max-pooled. All other parameters in the model are initialized randomly. The model is trained end-to-end for 15 epochs, with the Adam optimizer \citep{kingma2014adam}, sparse categorical cross-entropy loss, a learning rate of 0.001, and a minibatch size of 128. The best model is stored and the learning rate is reduced by a factor of 0.1 if validation loss does not decline after two successive epochs.

\section{Baseline Models}
The performance of the proposed model is compared with a host of machine learning and deep learning models and the results are reported in table~\ref{performance-comparison}. They are as follows:
\paragraph{Feature Based models:}
Multinomial Naive Bayes with bag-of-words input (MNB + BoW), Multinomial Naive Bayes with tf-idf input (MNB + TF-IDF), Linear SVC with bag-of-words input (LSVC + BoW), and Linear SVC with tf-idf input (LSVC + TF-IDF).
\paragraph{Basic Neural Networks:}
Feed forward Neural network with max-pooling (FFNN), CNN with max-pooling (CNN), and BiLSTM with maxpooling (BiLSTM)
\paragraph{Complex Neural Networks:}
BiLSTM +attention \citep{zhou2016attention} , serial BiLSTM-CNN \citep{chen2017improving}, and serial BiLSTM-CNN + attention.

\begin{table}
\centering
\begin{tabular}{lllll}
\hline \textbf{Metrics} & \textbf{bioche} & \textbf{com\_tech} & \textbf{cse} & \textbf{phy}\\ \hline
Precision & 0.9128 & 0.8831 & 0.9145 & 0.8931 \\
Recall & 0.7976 & 0.9342 & 0.8949 & 0.8793 \\
F1-Score & 0.8513 & 0.9079 & 0.9046 & 0.8862 \\
\hline
\end{tabular}
\caption{\label{model-performance} Detailed performance of the proposed
model on the validation data. }
\end{table}

\section{Results and Discussion}
The performance of all the models is listed in Table~\ref{performance-comparison}. The proposed model outperforms all other models in validation accuracy and weighted f1-score. It achieves better results than standalone CNN and BiLSTM, thus reasserting the importance of combining both the structures. The BiLSTM with attention model is similar to the proposed model, but the context is ignored. As the proposed model outperforms the BiLSTM with attention model, it proves the effectiveness of the CNN layer for providing context. Stacking a convolutional layer over a BiLSTM unit results in lower performance than the standalone BiLSTM. It can be thus inferred that combining CNN and BiLSTM in a parallel way is much more effective than just serially stacking. Thus, the attention mechanism proposed is able to successfully unify the CNN and the BiLSTM, providing meaningful context to the temporal representation generated by BiLSTM. Table~\ref{model-performance} reports the detailed performance of the proposed model for the validation data. The precision and recall for communication technology (com\_tech), computer science (cse), and physics(phy) labels are quite consistent. Biochemistry (bioche) label suffers from a high difference in precision and recall. This can be traced back to the fact that less amount of training data is available for the label, leading to the model overfitting on it.

\begin{table}
\centering
\begin{tabular}{lll}
\hline \textbf{Label} & \textbf{Validation} & \textbf{Validation} \\
& \textbf{Accuracy} &  \textbf{F1-Score} \\ \hline
MNB + Bow & 86.74 & 0.8352 \\
MNB + TF-IDF & 77.16 & 0.8010 \\
Linear SVC + Bow & 85.76 & 0.8432 \\
Linear SVC + TF-IDF & 88.17 & 0.8681 \\
FFNN & 76.11 & 0.7454 \\
CNN & 86.67 & 0.8532 \\
BiLSTM & 89.31 & 0.8842 \\
BiLSTM + Attention & 88.14 & 0.8697 \\ 
Serial BiLSTM-CNN & 88.99 & 0.8807 \\
Serial BiLSTM-CNN \\ + Attention & 88.23 & 0.8727 \\
\textbf{Ensemble CNN-BiLSTM} \\ \textbf{+ Attention} & \textbf{89.57} & \textbf{0.8875} \\ 
\hline
\end{tabular}
\caption{\label{performance-comparison} Performance comparison of different
models on the validation data. }
\end{table}

\section{Conclusion and Future work}

While NLP research in English is achieving new heights, the progress in low resource languages is still in its nascent stage. The TechDOfication task paves the way for research in this field through the task of technical domain identification for texts in Indian languages. This paper proposes a CNN-BiLSTM based attention ensemble model for the subtask-1f of Marathi text classification. The parallel CNN-BiLSTM attention-based model unifies the intermediate representations generated by both the models successfully using the attention mechanism. It provides a way for further research in adapting attention-based models for low resource and morphologically rich languages. The performance of the model can be enhanced by giving additional inputs such as character n-grams and document-topic distribution. More efficient attention mechanisms can be applied to further consolidate the amalgamation of CNN and RNN.

\bibliography{references}

\begin{thebibliography}{30}
\expandafter\ifx\csname natexlab\endcsname\relax\def\natexlab#1{#1}\fi

\bibitem[{Bojanowski et~al.(2017)Bojanowski, Grave, Joulin, and
  Mikolov}]{bojanowski2017enriching}
Piotr Bojanowski, Edouard Grave, Armand Joulin, and Tomas Mikolov. 2017.
\newblock Enriching word vectors with subword information.
\newblock \emph{Transactions of the Association for Computational Linguistics},
  5:135--146.

\bibitem[{Bolaj and Govilkar(2016{\natexlab{a}})}]{bolaj2016survey}
Pooja Bolaj and Sharvari Govilkar. 2016{\natexlab{a}}.
\newblock A survey on text categorization techniques for indian regional
  languages.
\newblock \emph{International Journal of computer science and Information
  Technologies}, 7(2):480--483.

\bibitem[{Bolaj and Govilkar(2016{\natexlab{b}})}]{bolaj2016text}
Pooja Bolaj and Sharvari Govilkar. 2016{\natexlab{b}}.
\newblock Text classification for marathi documents using supervised learning
  methods.
\newblock \emph{International Journal of Computer Applications}, 155(8):6--10.

\bibitem[{Chen et~al.(2017)Chen, Xu, He, and Wang}]{chen2017improving}
Tao Chen, Ruifeng Xu, Yulan He, and Xuan Wang. 2017.
\newblock Improving sentiment analysis via sentence type classification using
  bilstm-crf and cnn.
\newblock \emph{Expert Systems with Applications}, 72:221--230.

\bibitem[{Conneau et~al.(2016)Conneau, Schwenk, Barrault, and
  Lecun}]{conneau2016very}
Alexis Conneau, Holger Schwenk, Lo{\"\i}c Barrault, and Yann Lecun. 2016.
\newblock Very deep convolutional networks for text classification.
\newblock \emph{arXiv preprint arXiv:1606.01781}.

\bibitem[{Dhar et~al.(2018)Dhar, Mukherjee, Dash, and
  Roy}]{dhar2018performance}
Ankita Dhar, Himadri Mukherjee, Niladri~Sekhar Dash, and Kaushik Roy. 2018.
\newblock Performance of classifiers in bangla text categorization.
\newblock In \emph{2018 International Conference on Innovations in Science,
  Engineering and Technology (ICISET)}, pages 168--173. IEEE.

\bibitem[{Er et~al.(2016)Er, Zhang, Wang, and Pratama}]{er2016attention}
Meng~Joo Er, Yong Zhang, Ning Wang, and Mahardhika Pratama. 2016.
\newblock Attention pooling-based convolutional neural network for sentence
  modelling.
\newblock \emph{Information Sciences}, 373:388--403.

\bibitem[{Gao et~al.(2018)Gao, Ramanathan, and Tourassi}]{gao2018hierarchical}
Shang Gao, Arvind Ramanathan, and Georgia Tourassi. 2018.
\newblock Hierarchical convolutional attention networks for text
  classification.
\newblock In \emph{Proceedings of The Third Workshop on Representation Learning
  for NLP}, pages 11--23.

\bibitem[{Graves and Schmidhuber(2005)}]{graves2005framewise}
Alex Graves and J{\"u}rgen Schmidhuber. 2005.
\newblock Framewise phoneme classification with bidirectional lstm and other
  neural network architectures.
\newblock \emph{Neural networks}, 18(5-6):602--610.

\bibitem[{Guo et~al.(2018)Guo, Zhang, Wang, Wang, and Cui}]{guo2018cran}
Long Guo, Dongxiang Zhang, Lei Wang, Han Wang, and Bin Cui. 2018.
\newblock Cran: a hybrid cnn-rnn attention-based model for text classification.
\newblock In \emph{International Conference on Conceptual Modeling}, pages
  571--585. Springer.

\bibitem[{{Hassan} and {Mahmood}(2017)}]{8260793}
A.~{Hassan} and A.~{Mahmood}. 2017.
\newblock \href {https://doi.org/10.1109/ICMLA.2017.00009} {Efficient deep
  learning model for text classification based on recurrent and convolutional
  layers}.
\newblock In \emph{2017 16th IEEE International Conference on Machine Learning
  and Applications (ICMLA)}, pages 1108--1113.

\bibitem[{Hochreiter and Schmidhuber(1997)}]{hochreiter1997long}
Sepp Hochreiter and J{\"u}rgen Schmidhuber. 1997.
\newblock Long short-term memory.
\newblock \emph{Neural computation}, 9(8):1735--1780.

\bibitem[{Joshi et~al.(2019)Joshi, Goel, and Joshi}]{joshi2019deep}
Ramchandra Joshi, Purvi Goel, and Raviraj Joshi. 2019.
\newblock Deep learning for hindi text classification: A comparison.
\newblock In \emph{International Conference on Intelligent Human Computer
  Interaction}, pages 94--101. Springer.

\bibitem[{Kakwani et~al.(2020)Kakwani, Kunchukuttan, Golla, Gokul,
  Bhattacharyya, Khapra, and Kumar}]{kakwani2020inlpsuite}
Divyanshu Kakwani, Anoop Kunchukuttan, Satish Golla, NC~Gokul, Avik
  Bhattacharyya, Mitesh~M Khapra, and Pratyush Kumar. 2020.
\newblock inlpsuite: Monolingual corpora, evaluation benchmarks and pre-trained
  multilingual language models for indian languages.
\newblock In \emph{Proceedings of the 2020 Conference on Empirical Methods in
  Natural Language Processing: Findings}, pages 4948--4961.

\bibitem[{Kim(2014)}]{kim2014convolutional}
Yoon Kim. 2014.
\newblock Convolutional neural networks for sentence classification.
\newblock \emph{arXiv preprint arXiv:1408.5882}.

\bibitem[{Kingma and Ba(2014)}]{kingma2014adam}
Diederik~P Kingma and Jimmy Ba. 2014.
\newblock Adam: A method for stochastic optimization.
\newblock \emph{arXiv preprint arXiv:1412.6980}.

\bibitem[{Le et~al.(2017)Le, Cerisara, and Denis}]{le2017convolutional}
Hoa~T Le, Christophe Cerisara, and Alexandre Denis. 2017.
\newblock Do convolutional networks need to be deep for text classification?
\newblock \emph{arXiv preprint arXiv:1707.04108}.

\bibitem[{Liu et~al.(2016)Liu, Qiu, and Huang}]{liu2016recurrent}
Pengfei Liu, Xipeng Qiu, and Xuanjing Huang. 2016.
\newblock Recurrent neural network for text classification with multi-task
  learning.
\newblock \emph{arXiv preprint arXiv:1605.05101}.

\bibitem[{Liu et~al.(2020)Liu, Huang, Lu, and Lyu}]{liu2020multichannel}
Zhenyu Liu, Haiwei Huang, Chaohong Lu, and Shengfei Lyu. 2020.
\newblock Multichannel cnn with attention for text classification.
\newblock \emph{arXiv preprint arXiv:2006.16174}.

\bibitem[{Mikolov et~al.(2013)Mikolov, Chen, Corrado, and
  Dean}]{mikolov2013efficient}
Tomas Mikolov, Kai Chen, Greg Corrado, and Jeffrey Dean. 2013.
\newblock Efficient estimation of word representations in vector space.
\newblock \emph{arXiv preprint arXiv:1301.3781}.

\bibitem[{Pennington et~al.(2014)Pennington, Socher, and
  Manning}]{pennington2014glove}
Jeffrey Pennington, Richard Socher, and Christopher~D Manning. 2014.
\newblock Glove: Global vectors for word representation.
\newblock In \emph{Proceedings of the 2014 conference on empirical methods in
  natural language processing (EMNLP)}, pages 1532--1543.

\bibitem[{Sundermeyer et~al.(2015)Sundermeyer, Ney, and
  Schl{\"u}ter}]{sundermeyer2015feedforward}
Martin Sundermeyer, Hermann Ney, and Ralf Schl{\"u}ter. 2015.
\newblock From feedforward to recurrent lstm neural networks for language
  modeling.
\newblock \emph{IEEE/ACM Transactions on Audio, Speech, and Language
  Processing}, 23(3):517--529.

\bibitem[{Tummalapalli et~al.(2018)Tummalapalli, Chinnakotla, and
  Mamidi}]{tummalapalli2018towards}
Madhuri Tummalapalli, Manoj Chinnakotla, and Radhika Mamidi. 2018.
\newblock Towards better sentence classification for morphologically rich
  languages.
\newblock In \emph{Proceedings of the International Conference on Computational
  Linguistics and Intelligent Text Processing}.

\bibitem[{Vaswani et~al.(2017)Vaswani, Shazeer, Parmar, Uszkoreit, Jones,
  Gomez, Kaiser, and Polosukhin}]{vaswani2017attention}
Ashish Vaswani, Noam Shazeer, Niki Parmar, Jakob Uszkoreit, Llion Jones,
  Aidan~N Gomez, Lukasz Kaiser, and Illia Polosukhin. 2017.
\newblock Attention is all you need.
\newblock \emph{arXiv preprint arXiv:1706.03762}.

\bibitem[{Wang et~al.(2016)Wang, Huang, Zhu, and Zhao}]{wang2016attention}
Yequan Wang, Minlie Huang, Xiaoyan Zhu, and Li~Zhao. 2016.
\newblock Attention-based lstm for aspect-level sentiment classification.
\newblock In \emph{Proceedings of the 2016 conference on empirical methods in
  natural language processing}, pages 606--615.

\bibitem[{Xiao and Cho(2016)}]{xiao2016efficient}
Yijun Xiao and Kyunghyun Cho. 2016.
\newblock \href {http://arxiv.org/abs/1602.00367} {Efficient character-level
  document classification by combining convolution and recurrent layers}.

\bibitem[{Yang et~al.(2016)Yang, Yang, Dyer, He, Smola, and
  Hovy}]{yang2016hierarchical}
Zichao Yang, Diyi Yang, Chris Dyer, Xiaodong He, Alex Smola, and Eduard Hovy.
  2016.
\newblock Hierarchical attention networks for document classification.
\newblock In \emph{Proceedings of the 2016 conference of the North American
  chapter of the association for computational linguistics: human language
  technologies}, pages 1480--1489.

\bibitem[{Zhang et~al.(2015)Zhang, Zhao, and LeCun}]{zhang2015character}
Xiang Zhang, Junbo Zhao, and Yann LeCun. 2015.
\newblock Character-level convolutional networks for text classification.
\newblock In \emph{Advances in neural information processing systems}, pages
  649--657.

\bibitem[{Zheng and Zheng(2019)}]{zheng2019hybrid}
Jin Zheng and Limin Zheng. 2019.
\newblock A hybrid bidirectional recurrent convolutional neural network
  attention-based model for text classification.
\newblock \emph{IEEE Access}, 7:106673--106685.

\bibitem[{Zhou et~al.(2016)Zhou, Shi, Tian, Qi, Li, Hao, and
  Xu}]{zhou2016attention}
Peng Zhou, Wei Shi, Jun Tian, Zhenyu Qi, Bingchen Li, Hongwei Hao, and Bo~Xu.
  2016.
\newblock Attention-based bidirectional long short-term memory networks for
  relation classification.
\newblock In \emph{Proceedings of the 54th Annual Meeting of the Association
  for Computational Linguistics (Volume 2: Short Papers)}, pages 207--212.

\end{thebibliography}
\bibliographystyle{acl_natbib}

\end{document}